\documentclass[sigconf,nonacm]{acmart}

\usepackage{xcolor}
\usepackage{colortbl}
\usepackage{graphicx}
\usepackage{makecell}
\usepackage{tabularx}
\usepackage{multirow}
\usepackage{xspace}
\usepackage{enumitem}
\usepackage{float}
\usepackage{longtable}
\usepackage[most]{tcolorbox}
\usepackage{colortbl}
\usepackage{adjustbox}
\usepackage{pifont}
\usepackage{balance}
\usepackage{tablefootnote}

\usepackage{etoolbox}

\usepackage{titlesec} % 加载 titlesec 包
\titleformat{\section}
  {\normalfont\Large\bfseries}
  {\thesection}{1em}{\MakeUppercase} % 将 section 标题设置为全大写

\newcommand{\redcross}{\textcolor{red}{\ding{55}}}  % 红色叉
\newcommand{\bluetick}{\textcolor{blue}{\ding{51}}}  % 改为蓝色

\definecolor{mycolor}{RGB}{134,150,167}
\definecolor{backred}{RGB}{255, 190, 190}
\definecolor{backblue}{RGB}{210, 230, 250}
\definecolor{lbcolor}{RGB}{13, 151, 175}
\usepackage{color}

\definecolor{brown}{RGB}{139,64,0}
\AtBeginDocument{%
  }

\setcopyright{acmlicensed}
\copyrightyear{2018}
\acmYear{2018}
\acmDOI{XXXXXXX.XXXXXXX}
\acmConference[Conference acronym 'XX]{Make sure to enter the correct
  conference title from your rights confirmation email}{June 03--05,
  2018}{Woodstock, NY}
\acmISBN{978-1-4503-XXXX-X/2018/06}

\begin{document}

%%

\title{A Survey of WebAgents: Towards Next-Generation AI Agents for Web Automation with Large Foundation Models}


\author{Liangbo Ning$^1$, Ziran Liang$^1$, Zhuohang Jiang$^1$, Haohao Qu$^1$, Yujuan Ding$^1$, \\
Wenqi Fan$^1$*, Xiao-yong Wei$^1$, Shanru Lin$^2$, Hui Liu$^3$, Philip S. Yu$^4$, Qing Li$^1$}\authornote{Corresponding Authors: Wenqi Fan and Qing Li.}
\affiliation{%
  \institution{$^1$The Hong Kong Polytechnic University, $^2$City University of Hong Kong, \\$^3$Michigan State University, $^4$University of Illinois at Chicago}
  \country{}
}

\renewcommand{\shortauthors}{Liangbo Ning, Ziran Liang, Zhuohang Jiang, Haohao Qu, et al.}

\begin{abstract}
With the advancement of web techniques, they have significantly revolutionized various aspects of people's lives. Despite the importance of the web, many tasks performed on it are repetitive and time-consuming, negatively impacting overall quality of life. To efficiently handle these tedious daily tasks, one of the most promising approaches is to advance autonomous agents based on Artificial Intelligence (AI) techniques, referred to as \textbf{AI Agents}, as they can operate continuously without fatigue or performance degradation. In the context of the web, leveraging AI Agents -- termed \textbf{WebAgents} -- to automatically assist people in handling tedious daily tasks can dramatically enhance productivity and efficiency. Recently, Large Foundation Models (\textbf{LFMs}) containing billions of parameters have exhibited human-like language understanding and reasoning capabilities, showing proficiency in performing various complex tasks. This naturally raises the question: `\textit{Can LFMs be utilized to develop powerful AI Agents that automatically handle web tasks, providing significant convenience to users?}' 
To fully explore the potential of LFMs, extensive research has emerged on WebAgents designed to complete daily web tasks according to user instructions, significantly enhancing the convenience of daily human life. In this survey, we comprehensively review existing research studies on WebAgents across three key aspects: architectures, training, and trustworthiness. Additionally, several promising directions for future research are explored to provide deeper insights. 
\end{abstract}


\maketitle

\section{Introduction}

\begin{figure}
    \centering
    \includegraphics[width=0.95\linewidth]{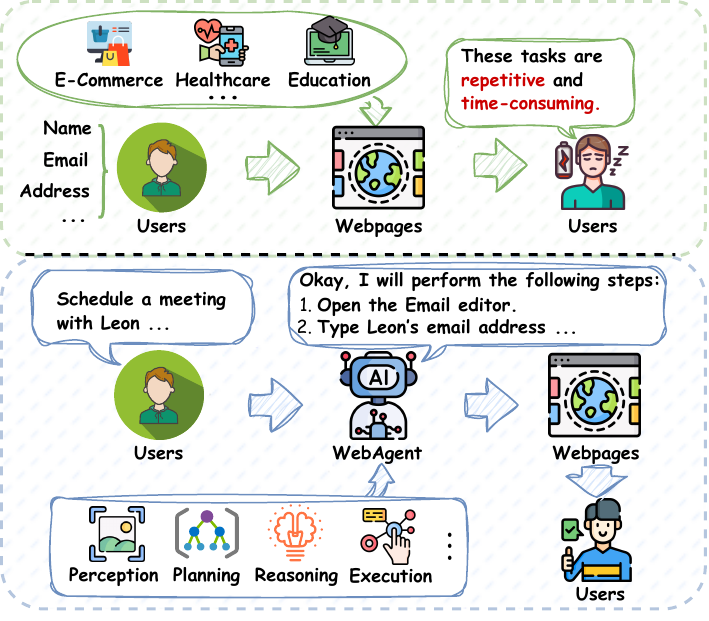}
    \vskip -0.15in
    \caption{Illustration of basic web tasks and the pipeline of WebAgents. Given the user instruction, WebAgents autonomously complete tasks by perceiving the environment, reasoning action sequences, and executing interactions. 
    }
    \label{fig:intro}
    \vskip -0.15in
\end{figure}

\noindent
``\textit{\textbf{The Web does not just connect machines, it connects people.}}''

\begin{flushright}
--- Tim Berners-Lee, \\
Inventor of the World Wide Web, \\
ACM A.M. Turing Award Laureate 2016.
\end{flushright}

\vskip 0.025in

As the web has rapidly evolved, it has profoundly transformed various aspects of people's lives, including information access~\cite{kobayashi2000information,kumar2017survey,fan2024survey}, shopping experiences~\cite{khalifa2003drivers,cai2016online}, and communications~\cite{carbone2012structured,ruiz2012survey}. 
For instance, the web serves as the largest knowledge repository to date, offering instant access to news~\cite{iglesias2016web,zhang2020overview}, academic papers (e.g., ArXiv~\cite{clement2019use}), and encyclopedias (e.g., Wikipedia~\cite{wu2007autonomously,volkel2006semantic}), enabling individuals to freely acquire desired information. 
This advancement has eliminated geographical barriers, providing people in remote areas with access to critical resources in education, healthcare, and law. 
Despite the importance of the web, many daily tasks we perform on it are repetitive and extremely time-consuming. 
For example, as shown in Figure~\ref{fig:intro}, when registering accounts on various platforms or filling out different application forms, we are often required to repeatedly enter the same personal information, such as our name, contact details, and address.
Similarly, when purchasing a product, we need to compare numerous options, review their ratings and prices, and ultimately decide on the final purchase. 
To effectively execute tedious daily tasks, one of the most promising techniques is to develop automatic agents embedded with human intelligence by taking advantage of Artificial Intelligence (AI) techniques, known as \textbf{AI Agents}.
In addition, AI Agents can execute tasks continuously without fatigue or performance degradation~\cite{hu2024agents}, ensuring reliability in repetitive workflows. 
Therefore, leveraging AI Agents -- \emph{termed \textbf{WebAgents} in the context of web} -- to assist people in handling tedious daily tasks automatically can extremely enhance productivity and efficiency, thereby further improving their quality of life.

Recently, large foundation models (\textbf{LFMs}) with billions of parameters, trained on massive data, have exhibited emergent human-like capabilities such as comprehension and reasoning, revolutionizing various domains including healthcare~\cite{fan2024survey,goyal2024healai,qiu2024llm}, E-commerce~\cite{fang2024llm,nie2024hybrid,wang2025knowledge}, and AI4Science~\cite{guo2024can,fan2025computational}.
For example, LFMs are integrated with protein data to capture the foundational protein knowledge, enabling better understanding and generation of protein structures, which can significantly advance the development of drug discovery and disease mechanism research~\cite{fan2025computational}. 
The human-like reasoning capabilities of LFMs are also leveraged in recommender systems (RecSys) to provide better item recommendations, significantly enhancing user online experience~\cite{zhao2024recommender,wu2024survey,huang2025towards}. 
By leveraging their extensive open-world knowledge, advanced instruction-following, and language comprehension and reasoning abilities, LFMs exhibit proficiency in simulating human-like behaviors to execute a variety of complex tasks.  
This naturally raises the promising topic: `\textit{Can LFMs be utilized to develop powerful AI Agents that automatically handle web tasks, providing significant convenience to users?}'

To fully explore the potential of LFMs, recent efforts have been made to advance LFMs-empowered \textbf{WebAgents} to complete various web tasks according to user instructions~\cite{hu2024agents}.  
For instance, the recent debut of a novel AI Agent framework named \emph{AutoGPT} has attracted significant interest from both academic and industrial communities, which exhibits impressive capabilities in autonomously handling complex tasks across both work and daily environments~\cite{Significant_Gravitas_AutoGPT}.
Unlike chatbots, AutoGPT can plan and execute complex tasks independently, performing automated searches and multi-step actions without requiring ongoing user instructions and supervision. 
In this context, as illustrated in Figure~\ref{fig:intro}, users only need to provide a natural language instruction, such as ‘\emph{Schedule a meeting with Leon at Starbucks on November 23, 2024, at 4:00 pm via email.}’ WebAgents can autonomously open the `Email' application, retrieve Leon’s email address, compose the email, and send it, thereby automating the entire scheduling process and greatly enhancing the convenience of daily life. 
Given the remarkable progress in developing LFM-empowered WebAgents and the growing number of related studies, there is a pressing need for a systematic review of recent advances in this field. 

To bridge this gap, this survey provides a comprehensive overview of WebAgents by summarizing representative methods from the perspectives of architecture, training, and trustworthiness. 
Specifically, Section~\ref{sec:preliminary} introduces the background knowledge of traditional agents and the fundamental pipeline of WebAgents. 
In Section~\ref{sec:target}, we review existing studies based on the three processes of WebAgents: perception, planning \& reasoning, and execution. 
Next, we summarize two crucial aspects (i.e., data and training strategies) in the training of WebAgents in Section~\ref{sec:training}. 
After that, we review studies that focus on investigating the trustworthy WebAgents, including their safety \& robustness, privacy, and generalizability, in Section~\ref{sec:trustworthy}. 
Finally, in Section~\ref{sec:future}, we discuss promising future research directions in WebAgents. 

Concurrently, several related surveys have been released, each exploring different aspects of autonomous agents. 
For example, \citet{hu2024agents} specifically review operating system agents, while \citet{huang2024understanding} primarily focus on the planning mechanisms of LFM-empowered agents.
\citet{xi2025rise} provide a broader perspective, offering a comprehensive overview of autonomous agents across various domains.
In contrast, our survey focuses specifically on WebAgents, systematically reviewing existing studies from three key perspectives: architectures, training, and trustworthiness.

\section{Background}
\label{sec:preliminary}
Recently, AI Agents with human-like reasoning and autonomous decision-making capabilities have revolutionized various domains, such as medicine\cite{li2024agent}, finance\cite{li2024personal}, and education\cite{jiang2024ai}. 
According to the foundational paradigm, existing AI Agents can be broadly divided into two categories: Reinforcement learning (RL)-based Agents and LFM-empowered Agents. 

\subsection{RL-based Agents}
In the early stage, numerous AI Agents leverage reinforcement learning to learn optimal policies by interacting with the environment and utilizing reward signals ~\cite{zhang2021multi}, achieving remarkable success in complex decision-making problems such as Go~\cite{silver2017mastering}, poker games~\cite{zhao2022alphaholdem}, robotic control~\cite{singh2022reinforcement}, and autonomous driving~\cite{kiran2021deep}. 
These RL-based agents primarily rely on value function optimization~\cite{littman2001value, taylor2009kernelized} and policy optimization~\cite{singh2000convergence, liu2021policy}. 
For example, 
Q-Learning~\cite{watkins1992q} is a value-iteration-based reinforcement learning algorithm that updates the Q-value function at each time step to evaluate the long-term reward of taking a specific action in a given state, and progressively approximates the optimal policy by leveraging a greedy strategy. 
Policy Gradient~\cite{sutton1999policy,fan2023adversarial,fan2021attacking} is a representative reinforcement learning approach that directly optimizes the policy by adjusting the parameters via gradient ascent to maximize the expected return, making it particularly suitable for continuous action spaces and high-dimensional complex tasks.

With the advancement of deep learning, deep neural networks (DNNs) have achieved remarkable success in fields such as computer vision~\cite{voulodimos2018deep} and data mining~\cite{wang2024multi,chen2022knowledge} by extracting and modeling complex representations from high-dimensional data. 
To harness the feature representation abilities of DNNs, extensive research has integrated them with reinforcement learning and proposed Deep Reinforcement Learning (DRL) algorithms~\cite{wang2022deep} for AI Agents. 
For example, AlphaGo~\cite{silver2017mastering} achieves superhuman performance in the game of Go by combining deep neural networks with Monte Carlo Tree Search (MCTS)~\cite{browne2012survey}, leveraging policy networks and value networks within deep reinforcement learning. 
Similarly, Financial Agent~\cite{hu2019deep} employs the deep recurrent neural network (RNN) model, simulating market environments and designing task-specific reward functions to achieve efficient and robust investment strategies in complex financial markets.

\subsection{LFM-empowered Agents}
In recent years, Large Foundation Models with billion-level parameters have demonstrated remarkable intelligence characterized by rich intrinsic knowledge~\cite{huang2024large,liu2024seamcarver}. 
Due to their human-like intelligence and rich open-world knowledge, leveraging LFMs as AI Agents has drawn considerable attention. 
For example, 
AIOS~\cite{mei2024aios} integrates large language models (LLMs) into operating systems to optimize resource allocation, enable context switching, and support concurrent agent execution. 
It also provides tool services and enforces access control, enhancing the performance and efficiency of LLM-based agents.  
CheatAgent~\cite{ning2024cheatagent} employs the LLMs as malicious agents to generate adversarial perturbations for misleading the victim LLM-based Recommender System (RecSys) by iteratively interacting with the target RecSys for policy optimization, thereby investigating their vulnerabilities. 
Besides,
LLMob~\cite{jiawei2024large} proposes an agent framework that integrates LLMs to simulate human behavior by combining habitual activity patterns and daily motivations, aiming to generate flexible and effective personal mobility trajectories.
Furthermore,
WorldCoder~\cite{tang2024worldcoder} introduces a world agent to dynamically generate and execute code to interact with environments, enabling iterative refinement of its internal capabilities for adapting to complex scenarios.

\subsection{AI Agents for Web Automation}
As LFM-based agents advance at an unprecedented pace, they are transforming how we interact with the digital world. 
Building on their remarkable success, LFM-based agents have been increasingly used for various web applications~\cite{zhou2023webarena,yang2024agentoccam,pan2024webcanvas}.
These LFM-empowered WebAgents aim to assist people in handling repetitive and time-consuming web tasks by emulating human behaviors. 
Mathematically, given a website $S$ (e.g., an online shopping site) and a user instruction $T$ (e.g., `Please help me buy a T-shirt'), WebAgents will generate a sequence of executable actions $\mathcal{A} = \{a_1, a_2, ..., a_n\}$ and interact with the environment based on these actions to complete the user task. 
Specifically, at the step $t$, WebAgents first observe the current environmental information $s_t$ from the screenshot or HTML of the website and retrieve the previous actions $\{a_1, a_2, ..., a_{t-1}\}$ as the short-term memory to guide the next-action prediction. 
After that, the user instruction $T$, observations $s_t$, and previous actions $\{a_1, a_2, ..., a_{t-1}\}$ are combined as the in-context knowledge to generate the next action $a_t$:  
$$a_t = f_{{\Theta}}(T, s_t, \{a_1, a_2, ..., a_{t-1}\}),$$
where $f_{{\Theta}}$ is the LFM-empowered WebAgents with parameters $\Theta$. 
Finally, WebAgents will emulate user behaviors and interact with the web environment based on the generated operation $a_t$. 
The state of the website will be updated to $s_{t+1}$ after executing the generated action, defined as:
$$s_{t+1} =  S(a_t).$$

\noindent
WebAgents will iteratively repeat the aforementioned steps until the user-given task is completed.

\section{WebAgent Architectures}\label{sec:target}

\begin{figure*}[htbp]
    \centering
    % \vskip -0.15in
    \includegraphics[width=0.97\linewidth]{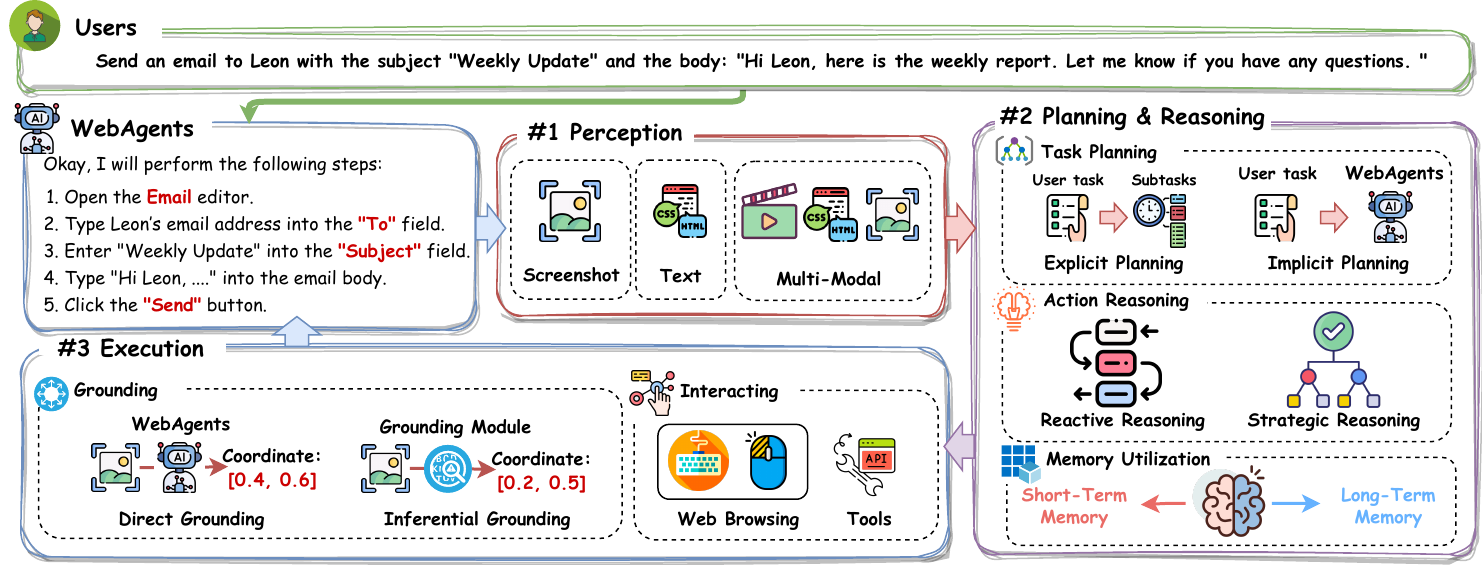}
    \vskip -0.15in
    \caption{Illustration of the overall framework of WebAgents, which contains three crucial processes: \textit{Perception}, \textit{Planning \& Reasoning}, and \textit{Execution}. Given the user's command, WebAgents first observe the environmental information during the perception process. Based on the observation, the action is generated in the planning and reasoning process. Finally, WebAgents execute the generated action to complete the user's task. }
    \label{fig:target}
    % \vskip -0.2in
\end{figure*}

\begin{table*}[!ht]
    \centering
    \scalebox{0.95}{
    \begin{tabular}{c|c|ccc|cc|c}
    \toprule
        \multirow{2}{*}{Model} & \multirow{2}{*}{Perception} & \multicolumn{3}{c|}{Planning \& Reasoning} & \multicolumn{2}{c|}{Execution} & \multirow{2}{*}{Date} \\ 
    \cline{3-7}
         &  & Task Planning & Action Reasoning & Memory Utilization & Interacting & Grounding & ~ \\ 
    \hline
        LCoW~\cite{lee2025learning} & TT     & \redcross & RR & STM & WB & IG & 03/2025 \\ 
        PLAN-AND-ACT~\cite{erdogan2025plan} & TT     & \bluetick & RR & STM & WB & DG & 03/2025 \\ 
        DoT~\cite{shao2025division} & - & \bluetick & RR & STM & - & - & 02/2025 \\ 
        COWPILOT~\cite{huq2025cowpilot} & - & \redcross & SR & STM & WB & - & 01/2025 \\ 
        R2D2~\cite{huang2025r2d2} & - & \redcross & SR & LTM & - & - & 01/2025 \\ 
        WEPO~\cite{liu2024wepo} & TT     & \redcross & RR & STM & WB & IG & 12/2024 \\ 
        Falcon-UI~\cite{shen2024falcon} & SS & \redcross & RR & STM & WB & DG & 12/2024 \\ 
        AGUVIS~\cite{xu2024aguvis} & SS & \redcross & RR & STM & WB & DG & 12/2024 \\ 
        Ponder \& Press~\cite{wang2024ponder} & SS & \redcross & RR & STM & WB & IG & 12/2024 \\ 
        ShowUI~\cite{lin2024showui} & SS & \redcross & SR & STM & WB & DG & 11/2024 \\ 
        WebDreamer~\cite{gu2024your} & TT & \redcross & SR & STM & WB & - & 11/2024 \\ 
        Auto-intent~\cite{kim2024auto} & TT     & \redcross & SR & STM & WB & DG & 10/2024 \\ 
        OSCAR~\cite{wang2024oscar} & SS & \bluetick & RR & STM & WB & IG & 10/2024 \\ 
        Infogent~\cite{reddy2024infogent} & - & \redcross & RR & STM & WB, TL     & - & 10/2024 \\ 
        API-calling agent~\cite{song2024beyond} & - & \redcross & RR & STM & WB, TL     & - & 10/2024 \\ 
        AgentOccam~\cite{yangagentoccam} & TT     & \bluetick & RR & STM & WB & - & 10/2024 \\ 
        Agent S~\cite{agasheagent} & MM & \bluetick & SR & LTM & WB & IG & 10/2024 \\ 
        ClickAgent~\cite{hoscilowicz2024clickagent} & SS & \redcross & RR & STM & WB & IG & 10/2024 \\ 
        D-PoT~\cite{zhang2024dynamic} & SS & \bluetick & RR & STM & WB & DG & 10/2024 \\ 
        NaviQAte~\cite{shahbandeh2024naviqate} & MM & \redcross & SR & LTM & WB & IG & 09/2024 \\ 
        AWM~\cite{wang2024agent} & - & \redcross & SR & LTM & - & - & 09/2024 \\ 
        Steward~\cite{tang2024steward} & SS & \redcross & RR & STM & WB & IG & 09/2024 \\ 
        WebPilot~\cite{zhang2024webpilot} & TT     & \bluetick & SR & STM & WB & - & 08/2024 \\ 
        AutoWebGLM~\cite{lai2024autowebglm} & MM & \redcross & RR & STM & WB & DG & 08/2024 \\ 
        OpenWebAgent~\cite{iong2024openwebagent} & MM & \redcross & RR & STM & WB & IG & 08/2024 \\ 
        OmniParser~\cite{lu2024omniparser} & SS & \redcross & - & - & - & IG & 08/2024 \\ 
        Agent-E~\cite{abuelsaad2024agent} & TT & \bluetick & RR & STM & WB & - & 07/2024 \\ 
        SearchAgent~\cite{koh2024tree} & SS & \redcross & SR & STM & - & - & 07/2024 \\ 
        CAAP~\cite{cho2024caap} & SS & \redcross & RR & STM & WB & - & 06/2024 \\ 
        MMAC-Copilot~\cite{song2024mmac} & MM & \bluetick & RR & STM & WB & - & 04/2024 \\ 
        % AIOS~\cite{mei2024aios} & - & - & - & - & TL & - & 03/2024 \\ 
        % Cradle~\cite{tan2024cradle} & SS & \bluetick & SR & LTM & WB & DG & 03/2024 \\ 
        OS-Copilot~\cite{wu2024copilot} & TT     & \bluetick & SR & LTM & WB, TL     & - & 02/2024 \\ 
        ScreenAgent~\cite{niu2024screenagent} & SS & \bluetick & RR & STM & WB & DG & 02/2024 \\ 
        UFO~\cite{zhang2024ufo} & SS & \bluetick & RR & LTM & WB & IG & 02/2024 \\ 
        DUAL-VCR~\cite{kil2024dual} & MM & \redcross & RR & STM & WB & IG & 02/2024 \\ 
        Seeclick~\cite{cheng2024seeclick} & SS & \redcross & RR & STM & WB & DG & 01/2024 \\ 
        WebVoyager~\cite{he2024webvoyager} & MM & \redcross & RR & STM & WB & IG & 01/2024 \\ 
        SeeAct~\cite{zheng2024gpt} & MM & \redcross & RR & STM & WB & IG & 01/2024 \\ 
        Assistgui~\cite{gao2023assistgui} & SS & \bluetick & RR & STM & WB & IG & 12/2023 \\ 
        LLMPA~\cite{guan2023intelligent} & SS & \bluetick & SR & STM & WB & IG & 12/2023 \\ 
        WebWISE~\cite{tao2023webwise} & TT     & \redcross & SR & LTM & WB & DG & 10/2023 \\ 
        ZSLA~\cite{li2023zero} & TT     & \bluetick & RR & STM & WB & - & 10/2023 \\ 
        Auto-GUI~\cite{zhang2023you} & SS & \redcross & RR & STM & WB & DG & 09/2023 \\ 
        % LASER~\cite{ma2023laser} & TT     & \redcross & SR & STM & WB & - & 09/2023 \\ 
        WebAgent~\cite{gur2024realworldwebagentplanninglong} & TT     & \bluetick & RR & STM & WB & - & 07/2023 \\ 
        Synapse~\cite{zheng2023synapse} & TT     & \redcross & SR & LTM & WB & - & 06/2023 \\ 
        MindAct~\cite{deng2023mind2web} & TT     & \redcross & RR & STM & WB & IG & 06/2023 \\ 
        ASH~\cite{lo2023hierarchical} & TT     & \redcross & RR & STM & WB & - & 05/2023 \\ 
        RCI~\cite{kim2023language} & TT     & \bluetick & SR & STM & WB & - & 03/2023 \\ 
    \bottomrule
    \end{tabular}
    }
    % \vskip -0.15in
    \caption{Basic publication information and main technical details of representative approaches. TT: Text-based WebAgents; SS: Screenshot-based WebAgents; MM: Multi-modal WebAgents; RR: Reactive Reasoning; SR: Strategic Reasoning; STM: Short-term Memory; LTM: Long-term Memory; DG: Direct Grounding; IG:  Inferential Grounding; WB: Web browsing-based WebAgents; TL: Tools-based WebAgents. `-' indicates that the article does not explicitly mention the technical details of this part.}
    \vskip -0.15in
    \label{tab:targets}
\end{table*}

There are three crucial and consecutive processes for WebAgents to fulfill user commands: 
\textbf{1) Perception} requires WebAgents to accurately observe the current environment, 
\textbf{2) Planning \& Reasoning} require WebAgents to analyze based on the current environment, interpret user-given tasks, and predict reasonable next actions, 
and \textbf{3) Execution} requires that WebAgents perform the generated actions and interact with the environment effectively. 
In the following section, we will comprehensively review the important techniques employed by WebAgents during these processes, and the details are also summarised in Table~\ref{tab:targets}. 

\subsection{Perception}
Typical large foundation models merely need to accept user instructions and generate corresponding responses through reasoning.
However, WebAgents, operating within complex web environments, are further expected to accurately perceive the external environment and perform behavioral reasoning based on the dynamic environment combined with the user's task. 
For instance, if a user requests WebAgents to open YouTube and play a video, WebAgents must first identify the location of the browser's address bar before they can proceed to input the URL \textit{`www.youtube.com'} and play the video. 
As shown in Figure~\ref{fig:target}, according to the data modality provided by the environment to WebAgents, we can categorize existing studies into three classes: 1) \textbf{Text-based}, 2) \textbf{Screenshot-based}, and 3) \textbf{Multi-modal} WebAgents. 

\subsubsection{Text-based WebAgents.}
With the advancement of large language models, extensive studies have been proposed to leverage its human-like understanding and reasoning abilities to assist users in addressing complex tasks. 
Since LLMs can only handle natural language, these WebAgents usually leverage the textual metadata of webpages (e.g., \textbf{HTML} and \textbf{accessibility trees}) to perceive the environment~\cite{kim2023language,lo2023hierarchical,zheng2023synapse,ma2023laser}. 
For example, MindAct~\cite{deng2023mind2web} introduces a two-stage framework that combines a fine-tuned small language model (LM) with an LLM to efficiently process large HTML documents, significantly reducing the input size while preserving essential information. This approach enables accurate prediction of both the target element and the corresponding action, effectively balancing efficiency and performance in web-based tasks. 
\citet{gur2024real} introduce an LLM-driven agent that learns from self-experience to complete tasks on real-world webpages. 
It summarizes long HTML documents into task-relevant snippets to extract the environmental information and decomposes user instructions into sub-tasks for effective planning.

\subsubsection{Screenshot-based WebAgents.}
Despite the remarkable success of text-based WebAgents, leveraging the textual metadata of the environment usually fails to align closely with human cognitive processes since the Graphical User Interfaces (GUI) are inherently visual~\cite{shen2024falcon,xu2024aguvis}. Additionally, textual representations usually vary across different environments and are verbose, leading to poor generalization abilities and increased computational overhead~\cite{zhang2024you}. 
Recently, breakthroughs in large vision-language models (VLMs) have significantly enhanced the capabilities of AI systems in processing complex visual interfaces. 
To leverage the visual understanding capabilities of VLMs, numerous studies have integrated them into WebAgents, utilizing screenshots to perceive the environment~\cite{zhang2024you,guan2023intelligent,gao2023assistgui,kil2024dual}. 
SeeClick~\cite{cheng2024seeclick} only relies on screenshots as observations to predict the next action and enhances the agent’s ability to locate relevant visual elements within screenshots by introducing a grounding pre-training process. 
OmniParser~\cite{lu2024omniparser} introduces an effective method to parse user interface screenshots into structured elements and enhances GPT-4V’s~\cite{yang2023dawn} ability to accurately ground actions to specific regions on the screen. 
OmniParser employs specialized models for parsing interactable regions on the screen and capturing the functions of detected elements, significantly improving the agent's performance on the visual benchmark.

\subsubsection{Multi-modal WebAgents.}
In addition to solely utilizing textual metadata or screenshots to comprehend the environment, numerous studies also leverage multi-modal data, combining their complementary strengths to provide WebAgents with a more comprehensive environmental perception~\cite{wang2024oscar,iong2024openwebagent,lai2024autowebglm}. 
For instance, 
MMAC-Copilot~\cite{song2024mmac} integrates GPT-4V for interpreting visual information from screenshots while leveraging Gemini Vision~\cite{li2024mini} to process and analyze video content, significantly enhancing the model's capabilities in handling multi-modal data. 
WebVoyager~\cite{he2024webvoyager} is a multi-modal WebAgent that autonomously completes web tasks end-to-end by processing both screenshots and textual content from interactive web elements. It leverages Set-of-Mark prompting~\cite{yang2023set} to overlay bounding boxes of the interactive elements on the webpages, significantly enhancing the agent's decision-making ability and enabling accurate action prediction and execution.

\subsection{Planning \& Reasoning}
Subsequent to the perception of environmental information, WebAgents are generally tasked with determining the appropriate action to execute the user's command. This involves analyzing the current state of the environment and utilizing the reasoning capabilities of LFMs.
As shown in Figure~\ref{fig:target}, there are three sub-tasks involved in this process: 
1) \textbf{Task Planning}, which focuses on reorganizing the user’s instruction and setting sub-objectives to help WebAgents effectively handle complex user queries;
2) \textbf{Action Reasoning}, which guides WebAgents to generate appropriate actions to fulfill the user’s commands; 
and 3) \textbf{Memory Utilization}, which equips WebAgents with internal information (e.g., previous actions) or external information (e.g., open-world knowledge from web search) to predict more appropriate actions.

\subsubsection{Task Planning.}
In the context of WebAgents, the objective of task planning is to determine a sequence of steps that the agent should take to complete the user-defined task efficiently and effectively~\cite{xi2025rise}. 
Based on whether WebAgents explicitly involve a task decomposition process, existing studies can be categorized into two types: 1) \textbf{Explicit Planning} and 2) \textbf{Implicit Planning}. 
Explicit planning methods usually decompose user instructions into multiple sub-tasks and generate actions to complete them step by step~\cite{li2023zero,zhang2024ufo,tan2024cradle}. 
For example, 
ScreenAgent~\cite{niu2024screenagent} introduces a structured workflow where the agent decomposes user tasks into sub-tasks, describes the screenshot, and generates the next actions in a function-call format. 
It further incorporates a reflection phase, which guides the agent to decide whether to proceed, retry, or reformulate the plan based on the current progress, making the entire workflow aligned with human thought processes. 
OS-Copilot~\cite{wu2024copilot} introduces a versatile planner that decomposes complex user requests into manageable sub-tasks while retrieving relevant external information (e.g., available tools and operating system information) to assist planning. It employs LLMs to formalize the plan into a directed acyclic graph, enabling parallel execution of independent tasks to enhance efficiency in real-world multi-task scenarios.

Implicit planning WebAgents directly feed the user instruction and the environmental observation into agents without an explicit task decomposition process~\cite{ma2023laser,cho2024caap,koh2024tree}. 
For instance, 
WebWISE~\cite{tao2023webwise} directly provides the task message to the agent and prompts it to generate actions progressively based on filtered Document Object Model (DOM) elements as environmental observations. 
OpenWebAgent~\cite{iong2024openwebagent} first uses a Web Processing Module to extract and simplify HTML elements, perform Optical Character Recognition (OCR)~\cite{singh2012survey} on screenshots, and annotate interactive elements. Then, OpenWebAgent introduces an Action Generation Module to predict the next action by directly feeding the current task and webpage context into agents.

\subsubsection{Action Reasoning.}
During the process of completing user-defined tasks, action reasoning is one of the most critical steps. It involves leveraging the agent's reasoning capabilities and current environmental observations to infer the next action. 
Based on the strategies, existing reasoning methods can be generally divided into two categories: 1) \textbf{Reactive Reasoning} and 2) \textbf{Strategic Reasoning}. 
Reactive reasoning means that WebAgents simply receive input prompts, including observations and instructions, and directly generate the next actions without additional operations~\cite{tang2024steward,zhang2024dynamic,hoscilowicz2024clickagent,zhang2024webpilot}. 
For instance, 
Agent-E~\cite{abuelsaad2024agent} introduces a planner agent to generate the next action without using complicated mechanisms during the action generation process. 
After that, a browser navigation agent equipped with multiple predefined foundational skills is employed to observe and interact with the environment to accomplish the user task. 
ASH~\cite{lo2023hierarchical} first proposes a SUMMARIZER prompt to condense the full webpage into an action-aware observation. During the action reasoning process, ASH directly utilizes an ACTOR prompt to predict the next action based on the summarized observation that contains relevant elements, improving efficiency and decision quality.

Strategic reasoning usually introduces additional operations to enhance the agent's action reasoning capability. Two of the most common approaches include incorporating an additional \textbf{exploration process} and integrating extra \textbf{in-context information}, both of which can effectively improve the accuracy of the agent's action generation~\cite{agasheagent,lin2024showui}. 
For example, 
WebDreamer~\cite{gu2024your} proposes a novel LLM-driven exploration strategy to simulate and predict the outcomes of candidate actions before execution using natural language descriptions, allowing the agent to evaluate and select the optimal action at each step. This action simulation mechanism can significantly enhance decision accuracy and efficiency while reducing unnecessary interactions with the webpage. 
Auto-intent~\cite{kim2024auto} introduces an unsupervised method to extract compact intents from multiple target domain demonstrations and trains an intent predictor to forecast the next intent based on past observations and actions. 
By providing top-k predicted intents as additional in-context hints to agents during action generation, Auto-intent enhances the agent’s performance on large-scale real-website navigation benchmarks.

\subsubsection{Memory Utilization.}
Beyond task planning and action reasoning, the effective utilization of memory is another key factor contributing to the powerful capabilities of WebAgents. 
Depending on its source, memory can generally be divided into 1) \textbf{Short-term Memory} and 2) \textbf{Long-term Memory}. 
Short-term memory generally denotes the previous actions performed to accomplish the user's task. Considering short-term memory when generating the next action can effectively avoid redundant operations and improve task completion efficiency~\cite{reddy2024infogent,hoscilowicz2024clickagent,zhang2024dynamic,lu2024omniparser}. 
For example, 
AutoWebGLM~\cite{lai2024autowebglm} formulates web browsing tasks as a sequential decision-making process, where the agent determines actions like clicking, scrolling, or typing based on the current state and history, including previous webpage states and their corresponding actions. 
LLMPA~\cite{guan2023intelligent} introduces a Previous Action Description Generator to generate action descriptions based on the high-level task description and the corresponding historical behavior sequences. Previous actions and their generated descriptions are injected into input prompts as the short-term memory for the next action prediction.

Long-term memory refers to external information that persists over time, such as action trajectories from previously executed tasks and knowledge acquired by online search. By retrieving this external knowledge as a reference, task success rates of WebAgents can be significantly enhanced~\cite{wang2024agent,tao2023webwise,zhang2024ufo,wu2024copilot}.
For instance, 
Agent S~\cite{agasheagent} leverages both Online Web Search for external knowledge and Narrative Memory for internal task-specific experience that includes summaries of both successful and failed trajectories to generate a sequence of sub-tasks that can accomplish the user instruction. After that, some similar sub-task experience is also retrieved and used for the action generator to predict the next action. 
Synapse~\cite{zheng2023synapse} first processes the raw web state into concise task-relevant observations to reduce the computational overhead and introduces a Trajectory-as-Exemplar (TaE) prompting strategy, which retrieves similar trajectories (i.e., sequences of abstracted states and actions) to guide the generation of the next action.

\subsection{Execution}
The final step for WebAgents to complete the user's command is to interact with the webpages and execute the generated actions. 
As shown in Figure~\ref{fig:target}, there are two tasks during this process: 1) \textbf{Grounding}, which aims to locate the elements within the webpage that agents will interact with, and 2) \textbf{Interacting}, which focuses on performing specific actions on the selected elements.

\subsubsection{Grounding.}
Since webpages often contain numerous interactive elements, selecting the correct element to execute the generated action is crucial for completing the user's task. 
Based on the grounding strategies of WebAgents, we categorize existing research into two categories: 1) \textbf{Direct Grounding} and 2) \textbf{Inferential Grounding}. 
Direct grounding refers to WebAgents directly generating the coordinates of the candidate element within a screenshot or selecting an element from the entire HTML for interactions~\cite{yangagentoccam,lai2024autowebglm,iong2024openwebagent,zhang2024dynamic}. 
For example, 
ShowUI~\cite{lin2024showui} directly guides the agent to generate the correct action (e.g., [CLICK]) and its corresponding parameters (e.g., coordinates for [CLICK]) to locate the elements within the webpage that the action will perform. 
Auto-intent~\cite{kim2024auto} leverages the HTML elements as the in-context example and guides the agents to correctly select the target element that the generated action will operate on.

Inferential grounding involves leveraging extra auxiliary modules to locate target elements~\cite{reddy2024infogent,agasheagent,hoscilowicz2024clickagent,shahbandeh2024naviqate,lee2025learning}.
For instance, 
Ponder \& Press~\cite{wang2024ponder} introduces a general-purpose Multi-modal LLM (MLLM) as an interpreter, responsible for translating high-level user instructions into detailed action descriptions, and a GUI-specific MLLM as a locator, which accurately identifies the target GUI elements within visual screenshots based on the generated action descriptions. 
OSCAR~\cite{wang2024oscar} introduces a dual-grounding observation approach that contains visual grounding and semantic grounding. 
Visual grounding uses Set-of-Mark prompting~\cite{yang2023set} with precise element bounding boxes extracted from the accessibility tree, and explicit semantic grounding annotates key elements with descriptive labels.  
This dual-grounding strategy significantly improves GUI understanding, enabling OSCAR to accurately locate and interact with GUI elements in complex and dynamic web environments. 
MindAct~\cite{deng2023mind2web} introduces a two-stage framework that first uses a fine-tuned small language model to rank and filter webpage elements, followed by an LLM that processes the selected elements to predict both the target element and the required action.

\subsubsection{Interacting.}
Finally, WebAgents need to interact with the target element using the generated actions. 
By repeating the above steps multiple times, WebAgents can complete the user's complex task. 
Based on the approach WebAgents interact with the webpages, existing studies can be generally categorized into two classes: 1) \textbf{Web Browsing-based} and 2) \textbf{Tool-based} methods. 
Web browsing-based methods utilize typical actions that humans employ when navigating websites, such as clicking, scrolling, and typing~\cite{shen2024falcon,xu2024aguvis,wang2024ponder,lin2024showui,hoscilowicz2024clickagent}. 
For example, 
NaviQAte~\cite{shahbandeh2024naviqate} define three widely-used actions [Click], [Type], and [Select] to perform the web application navigation task. 
AgentOccam~\cite{yangagentoccam} simplifies the action space for WebAgents by removing redundant, low-utility, and embodiment-dependent actions, while introducing high-level commands like [Note], [Stop], and so on, significantly enhancing the interacting efficiency of WebAgents.

Tool-based methods involve using additional tools, such as application programming interfaces (APIs), to interact with the webpages~\cite{song2024beyond,gu2024your,mei2024aios,wu2024copilot}. 
For instance, 
API-calling agent~\cite{song2024beyond} introduces a novel framework that expands the traditional WebAgent action space by incorporating direct API interactions, enabling agents to bypass GUI-based interactions entirely, improving efficiency and adaptability for real-world online tasks. 
Infogent~\cite{gu2024your} proposes to utilize external tools like Google Search API to navigate the web and a scraper to extract webpage contents for web information aggregation tasks.

\begin{figure*}
    \centering
    % \vskip -0.15in
    % \includegraphics[width=0.8\linewidth]{Image/TrainingData.pdf}
    \includegraphics[width=0.97\linewidth]{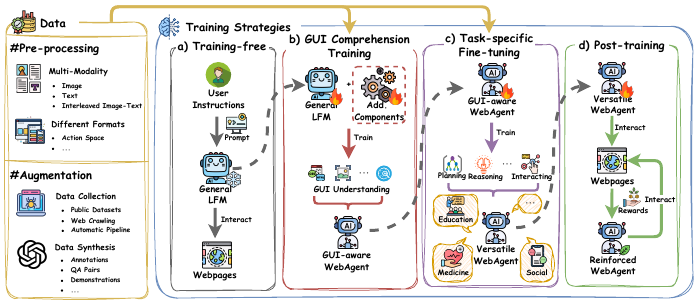}
    \vskip -0.15in
     \caption{
     Illustration of the training for WebAgents.
     There are two processes to construct the training data: 1) \textit{Data Pre-processing} aims to reduce discrepancy across various data modalities and formats, and 2) \textit{Data Augmentation} enhances the quantity and diversity of the training data.
     For training strategies, there are mainly four categories: 
     1) \textit{Training-free} approach directly utilizes prompts to guide LFMs in completing web tasks, 
     2) \textit{GUI Comprehension Training} aims to enhance GUI understanding capabilities of general-purpose LFMs, 
     3) \textit{Task-specific Fine-tuning} improves task-oriented capabilities of WebAgents, and 
     4) \textit{Post-training} interacts with webpages to receive rewards for further optimizing the policy of WebAgents.
     }
    \label{fig:trainingdata}
    % \vskip -0.2in
\end{figure*}

\section{Training of WebAgents}\label{sec:training}

There are two fundamental aspects in the training of WebAgents:
1) \textbf{Data} serves as the cornerstone of WebAgent training, providing diverse and representative samples that help models learn web-specific patterns.
2) \textbf{Training Strategies} indicate how WebAgents acquire and refine their capabilities through different approaches.
The following section provides a comprehensive review of existing studies, focusing on their methods for constructing training data and training strategies.
The overall training framework of WebAgents is presented in Figure~\ref{fig:trainingdata} and the details of representative works are summarized in Table~\ref{tab:trainingdata}.

\begin{table*}[!ht]
\centering
\scalebox{0.95}{
\begin{tabular}{cccccccc}
\toprule
\multicolumn{1}{c}{\multirow{3}{*}{Model}}         & \multicolumn{2}{c}{{{\raisebox{-0.14\height}{\includegraphics[height=1.3em]{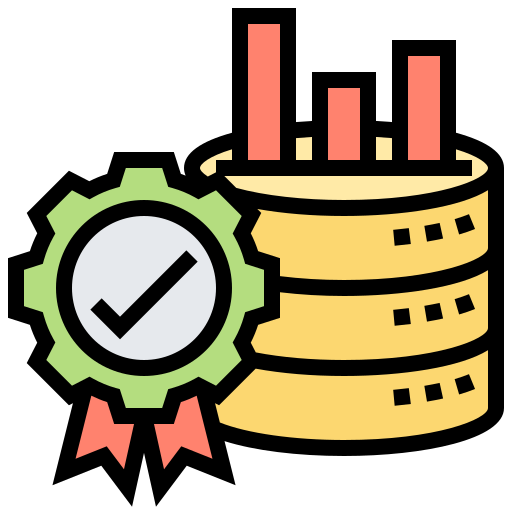}}}\hspace{0.5em}{Data}}}         & \multicolumn{4}{c}{{{\raisebox{-0.14\height}{\includegraphics[height=1.3em]{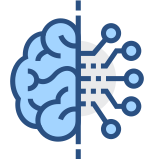}}}\hspace{0.5em}{Training Strategies}}}            &        \\
\cmidrule(lr){2-3} 
\cmidrule(lr){4-7} 
\multicolumn{1}{c}{}        &  \multirow{2}{*}{Pre-processing}      &  \multirow{2}{*}{Augmentation}  & \multirow{2}{*}{Training-free}            & GUI Comprehension  & Task-specific  & \multirow{2}{*}{Post-training }           &     Date       \\
\multicolumn{1}{c}{}        &      &    &        & Training &  Fine-tuning &          &            \\
\midrule

PLAN-AND-ACT~\cite{erdogan2025plan} &   \redcross & DS  & \redcross &   \redcross & \bluetick & \redcross & 03/2025 \\
LCoW~\cite{lee2025learning} & \redcross  & DC &    \redcross &   \redcross & \bluetick & \redcross & 03/2025 \\
ScreenSpot-Pro~\cite{li2025screenspot} &  \redcross &   \redcross  & \redcross &  \bluetick  & \redcross & \redcross & 03/2025 \\
% Division-of-Thoughts~\cite{shao2025division} &   &     &  &    &  & \bluetick & 6/02/2025 \\
NNetNav~\cite{murty2024nnetscape}           & \bluetick  & DS    & \redcross & \redcross   & \bluetick  & \redcross & 02/2025 \\
Iris~\cite{ge2024iris} & \redcross  &   \redcross  & \redcross &  \bluetick  & \redcross & \redcross & 02/2025 \\
% SHARPIE~\cite{aydin2025sharpie} &   &     &  &    &  & \bluetick & 3/02/2025 \\
Layered-CoT~\cite{sanwal2025layered} &  \redcross &  \redcross   & \bluetick &   \redcross & \redcross & \redcross & 01/2025 \\
AgentRefine~\cite{fu2025agentrefine} &  \bluetick &  DS   & \redcross &   \redcross & \bluetick & \redcross & 01/2025 \\
HTML-T5++~\cite{furuta2023exposing}         & -  &    -   & \redcross & \redcross   & \bluetick  & \redcross & 12/2024 \\
WEPO~\cite{liu2024wepo} & \redcross  &   \redcross  & \redcross &  \redcross  & \redcross  & \bluetick & 12/2024 \\
Falcon-UI~\cite{shen2024falcon}         &  \bluetick  & DC    & \redcross & \bluetick   & \redcross  & \redcross & 12/2024 \\
AgentTrek~\cite{xu2024agenttrek}         & \bluetick & DS    & \redcross & \bluetick   & \bluetick  & \redcross & 12/2024 \\
% AutoPrep~\cite{fan2024autoprep} &   &     &  &    &  &  & 10/12/2024 \\
Aguvis~\cite{xu2024aguvis}            & \bluetick & DC \& DS  & \redcross & \bluetick   & \bluetick  & \redcross & 12/2024 \\
ScribeAgent~\cite{shen2024scribeagent}       &  \bluetick & DC    & \redcross & \redcross   & \bluetick  & \redcross & 12/2024 \\
Synatra~\cite{ou2024synatra}           &  \bluetick & DS    & \redcross & \redcross   & \bluetick  & \redcross & 11/2024 \\
ShowUI~\cite{lin2024showui}            & \bluetick & DC \& DS  & \redcross & \bluetick   & \bluetick  & \redcross & 11/2024 \\
MultiUI~\cite{liu2024harnessing}           & \bluetick & DC    & \redcross & \bluetick   & \redcross  & \redcross & 11/2024 \\
EDGE~\cite{chen2024edge} & \bluetick  & DS    & \redcross & \bluetick   & \redcross  & \redcross & 11/2024 \\
OS-Atlas~\cite{wu2024atlas}          & \bluetick & DC \& DS & \redcross & \bluetick   & \redcross  & \redcross & 10/2024 \\
AutoGLM~\cite{liu2024autoglm}           & \redcross & \redcross     & \redcross & \bluetick   & \bluetick  & \bluetick & 10/2024 \\
Ferret-UI 2~\cite{li2024ferret}       &  \bluetick & DS    & \redcross & \bluetick   & \redcross  & \redcross & 10/2024 \\
AutoWebGLM~\cite{lai2024autowebglm}        &  \redcross & DC    & \redcross & \bluetick   & \bluetick  & \bluetick & 10/2024 \\
SeeAct-V~\cite{gou2024navigating}          & \bluetick & DS    & \redcross & \bluetick   & \redcross  & \redcross & 10/2024 \\
MobileVLM~\cite{wu2024mobilevlm}         &  \bluetick & DC    & \redcross & \bluetick   & \redcross  & \redcross & 10/2024 \\
MM1.5~\cite{zhang2024mm1}& \bluetick & \redcross     & \redcross & \bluetick   & \bluetick  & \redcross & 09/2024 \\
Textual Foresight~\cite{burns2024tell} & \bluetick &   DC \& DS    & \redcross & \bluetick   & \redcross  & \redcross & 08/2024 \\
CoAT~\cite{zhang2024android} & \bluetick & DC \& DS    & \bluetick & \redcross   & \redcross  & \redcross & 07/2024 \\
ScreenAI~\cite{baechler2024screenai}          & \bluetick & DS    & \redcross & \bluetick   & \bluetick  & \redcross & 07/2024 \\
Auto-GUI~\cite{zhang2023you}          &  \redcross &   \redcross    & \bluetick & \redcross   & \redcross  & \redcross & 06/2024 \\
LVG~\cite{qian2024visual} & \bluetick & DS    & \redcross & \bluetick   & \redcross  & \redcross & 06/2024 \\
UINav~\cite{li2023uinav} & \redcross & DC    & \redcross & \redcross   & \bluetick  & \redcross & 06/2024 \\
GLAINTEL~\cite{fereidouni2024grounded}          & \redcross & \redcross     & \redcross & \redcross   & \bluetick  & \bluetick & 04/2024 \\
\citet{dorka2024training}        & \redcross & \redcross     & \redcross & \redcross   & \bluetick  & \redcross & 04/2024 \\
\citet{thil2024navigating} & \bluetick & \redcross     & \redcross & \bluetick   & \bluetick  & \bluetick & 04/2024 \\
MM1\cite{mckinzie2024mm1} & \bluetick & \redcross     & \redcross & \bluetick   & \bluetick  & \redcross & 04/2024 \\
Ferret-UI~\cite{li2024ferret}       &  \bluetick & DC \& DS    & \redcross & \bluetick   & \bluetick  & \redcross & 04/2024 \\
WebAgent~\cite{gur2024realworldwebagentplanninglong}& \redcross & \redcross     & \redcross & \bluetick   & \bluetick  & \redcross & 02/2024 \\
WebGUM~\cite{furuta2023multimodal}            & \bluetick & DS    & \redcross & \bluetick   & \bluetick  & \redcross & 02/2024 \\
WebVLN~\cite{chen2024webvln}            &  \bluetick  & DC \& DS    & \redcross & \redcross   & \bluetick  & \redcross & 12/2023 \\
Pix2Struct~\cite{lee2023pix2struct}         & \redcross & \redcross    & \redcross & \bluetick   & \redcross  & \redcross & 12/2023 \\
LLMPA~\cite{guan2023intelligent} & \redcross & \redcross    & \redcross & \redcross   & \bluetick  & \redcross & 12/2023 \\
ILuvUI~\cite{jiang2023iluvui}            & \bluetick & DS    & \redcross & \bluetick   & \redcross  & \redcross & 10/2023 \\
RUIG~\cite{zhang2023reinforced} & \redcross & \redcross    & \redcross & \bluetick   & \redcross  & \bluetick & 10/2023 \\
Spotlight~\cite{li2022spotlight}         & \redcross & \redcross     & \redcross & \bluetick   & \redcross  & \redcross & 02/2023 \\
Lexi~\cite{banerjee2023lexi} & \bluetick & DC    & \redcross & \bluetick   & \redcross  & \redcross & 01/2023 \\
CC-Net~\cite{humphreys2022data}        &  \redcross &  DC     & \redcross  &   \redcross  &  \bluetick  & \bluetick & 11/2022\\
\bottomrule
\end{tabular}
}
\caption{
Basic publication information and main training details of representative WebAgents. In the Augmentation column, DC: Data Collection; DS: Data Synthesis. For papers proposing benchmarks, the Data column is labeled as `-' since this section focuses on training data only.
}
\label{tab:trainingdata}
\vskip -0.2in
\end{table*}

\subsection{Data}
Data serves as the foundation that powers contemporary WebAgents, with training data construction encompassing two essential steps: \textbf{1) Data Pre-processing}, which refines and structures the data to enhance its usability and quality, and \textbf{2) Data Augmentation}, which augments both the quantity and diversity of the dataset.

\subsubsection{Data Pre-processing.}
Data pre-processing is a critical step in ensuring data quality and suitability for training, including filtering~\cite{ou2024synatra}, curation~\cite{zhang2024ui}, mixture~\cite{zhang2024mm1}, and so on.
Since web environments typically feature multiple modalities, such as text and images, alongside diverse data formats across platforms, this data pre-processing process crucially relies on \textbf{Modality Alignment} and \textbf{Format Alignment} to address the unique challenges of heterogeneity and ensure effective data integration.
Specifically, modality alignment tackles the challenges of integrating multi-modal web data with discrepancies across modalities and inconsistent granularity, enabling models to harness their complementary strengths for a deeper understanding of the environment~\cite{burns2024tell}.
For instance,
\citet{liu2024harnessing} capture critical web elements and layout structures through screenshots and augmented accessibility trees, enhancing rich interactions between text and visuals while filtering out irrelevant data within websites.
\citet{gou2024navigating} construct <\textit{screenshot, referring expression, coordinates}> triplets for multi-modality alignment, where the referring expressions are generated based on HTML, incorporating visual, positional, and functional information.
LVG~\cite{qian2024visual} pairs UI screenshots with free-form language expressions through contrastive learning, whereas ILuvUI~\cite{jiang2023iluvui} combines pixel-based CLIP-L-336px~\cite{radford2021learning} with a generative LLM to produce paired text-image training data.
In addition, format alignment addresses platform-specific discrepancies, such as naming conflicts, ensuring consistent and effective data integration across diverse sources.
For example, the action \textit{tap} on mobile devices corresponds to \textit{click} on PCs, potentially confusing model consistency in understanding and executing tasks across platforms.
To address this issue, OS-Atlas~\cite{wu2024atlas} aligns the action space in cross-platform datasets, mitigating potential inconsistencies introduced by heterogeneous data integration.

\subsubsection{Data Augmentation.}\label{sec:data_aug}
Large-scale data serves as the foundation for the emergent intelligence of LFMs.
For LFM-empowered WebAgents, it is also crucial to collect large volumes of diverse training data, thereby enhancing their capabilities to accurately perceive website environments, reason next action, and execute complex web-based operations.
Depending on the methods of data acquisition, data augmentation approaches can be categorized into two distinct types: 1) \textbf{Data Collection} and 2) \textbf{Data Synthesis}.

Data collection involves gathering data from public datasets or directly from real-world scenarios.
For instance, Lexi~\cite{banerjee2023lexi} collects and curates 114k UI images paired with functional captions sourced from open websites, exhibiting remarkable diversity across applications, platforms, and UI characteristics.
Rather than indiscriminately aggregating all available data, some studies like ShowUI~\cite{lin2024showui} demonstrate that cautiously sampling high-quality, representative data from public datasets can significantly enhance WebAgents' performance.
While data annotated by human experts exhibits high quality, its labor-intensive process limits the total amount of the resulting datasets, leading to undertrained models with limited generalization to unseen online contexts.
To address this challenge, several autonomous generation methods have been proposed to construct cost-effective datasets without reliance on human assistance or large vision-language models~\cite{banerjee2023lexi,lee2025learning}.
For instance,
Falcon-UI~\cite{shen2024falcon} autonomously generates a dataset, Insight-UI, by interacting with publicly available webpages, comprising multi-step, cross-platform screenshots with all visible interactive elements and character-level OCR annotations.
ScribeAgent~\cite{shen2024scribeagent} collects action sequence data across various websites executed by real users from \href{https://scribehow.com}{\textit{Scribe}}. These sequences are structured as document-based workflows, encompassing a diverse range of targets and difficulty levels.
Besides, UINav~\cite{li2023uinav} mitigates the pressure of scarce training data by randomizing the attributes of secondary UI elements, such as text embedding or element offsets, to augment demonstrations without additional data collection overhead.

Data synthesis involves automatically generating plausible web-relevant datasets using LLMs or VLMs to enrich training data when real-world samples are insufficient or costly to obtain.
Beyond some basic data synthesis approaches that merely crawl elements from webpages and generate annotations for these elements, a group of studies focuses on creating rich question-answering (QA) pairs to further advance the GUI understanding of WebAgents, including general QA pairs~\cite{baechler2024screenai,chen2024webvln} and interaction-focused QA pairs~\cite{chen2024edge,li2024ferret}.
Complementing these QA-based approaches, some studies utilize VLMs to produce trajectory-related data that captures sequential interactions.
For example, AgentTrek~\cite{xu2024agenttrek} leverages collected web tutorials as step-by-step instructions to simulate execution trajectories, while Aguvis~\cite{xu2024aguvis} generates inner monologues of each step for existing trajectories.
Moreover, NNetNav~\cite{murty2024nnetscape} generates substantial, realistic, and interactive trajectory data through a hierarchical exploration policy and pruning techniques to navigate websites within an exponentially large space.
Besides, Synatra~\cite{ou2024synatra} transforms three types of indirect knowledge from web environments, including procedural knowledge, environment knowledge, and ungrounded observations, into direct demonstrations at scale to increase the richness of information in training data.

\subsection{Training Strategies}

Once large datasets are collected, effectively using them for WebAgent training is essential. 
Depending on the training strategies employed, which differ in learning paradigms, data usage, and optimization objectives, existing studies can generally be divided into four categories: 1) \textbf{Training-free}, 2) \textbf{GUI Comprehension Training}, 3) \textbf{Task-specific Fine-tuning}, and 4) \textbf{Post-training}.

\subsubsection{Training-free.}
The rapid evolution of LFMs, endowed with human-like intelligence and robust comprehension of visual and textual information, substantially boosts the development of intelligent WebAgents. 
Building upon these capabilities, training-free methods directly adapt LFMs as specialized WebAgents, harnessing well-crafted prompts to guide models in executing web tasks without any architectural modifications or parameter updates~\cite{chen2024comm,rawles2023androidinthewild,wei2022chain,wen2023empowering}.
For example,
Auto-GUI~\cite{zhang2023you} synthesizes both previous action histories and future action plans within a chain-format prompt to inform the decision-making process for determining the subsequent action step.
Advancing this methodology, 
CoAT~\cite{zhang2024android} proposes a chain-of-action-thought prompting paradigm, which integrates actions and thoughts to enable more efficient navigation capabilities.
% The paradigm incorporates screen descriptions, previous actions, and their outcomes to infer the reasoning behind the next action determination, alongside generating textual descriptions of the subsequent step and its potential results.
The paradigm incorporates screen descriptions, previous actions, and their current outcomes to offer explicit explanations behind the next action determination, alongside generating textual descriptions of the subsequent step and its potential results.

\subsubsection{GUI Comprehension Training.}
Although general-purpose LFMs are extensively trained on enormous datasets, these models often fall short in GUI-aware capabilities, particularly screen understanding and OCR, hindering them to comprehend and interact effectively with webpages. 
Specifically, general LFMs might focus on decorative icons or background text instead of the key interface elements, leading to misinterpretations of available functionalities~\cite{liu2024harnessing}.
To bridge this gap, numerous GUI comprehension training approaches are developed to enhance the critical foundational GUI understanding capabilities of WebAgents through supervised learning on large-scale web datasets~\cite{shen2024falcon,wu2024mobilevlm,dorka2024training,banerjee2023lexi}. 
For instance,
Aguvis~\cite{xu2024aguvis} introduces a two-stage training paradigm that begins with unifying GUI environments as images and concentrates on equipping the model to comprehend and interact with objects within a single GUI screenshot during the pre-training stage, laying a robust GUI comprehension foundation for subsequent fine-tuning.
Similarly, OS-Atlas~\cite{wu2024atlas} leverages vast and diverse high-quality triplets <\textit{screenshot, referring expression, coordinates}> to train an existing VLM by predicting the element coordinates with screenshots and referring expression, which enables the model to effectively interpret GUI screenshots and accurately identify their constituent elements and locations.
While Aguvis and OS-Atlas focus primarily on vision-centric GUI understanding learning, MM1.5~\cite{zhang2024mm1} capitalizes on text-rich OCR data during training to significantly enhance its ability to comprehend interleaved image-text data.
This capability is highly required in text-rich scenarios like E-commerce platforms, where product webpages often contain a large amount of detailed text descriptions, such as product specifications and reviews, as well as numerous images, like product photos and size charts.
In addition, given the context sensitivity of the UI elements, LVG~\cite{qian2024visual} introduces layout-guided contrastive learning, which captures the semantics of individual UI elements based on their visual organization, further enhancing the model's GUI comprehension.

Beyond the aforementioned approaches that focus on introducing new training objectives or stages, several approaches also incorporate additional components during training to strengthen the GUI understanding capabilities of general-purpose LFMs. 
For example,
Spotlight~\cite{li2022spotlight} implements Region Summarizer to extract the most essential regions of screenshots based on VLM encoding, thereby facilitating GUI understanding.
ScreenAI~\cite{baechler2024screenai} employs Pix2Struct~\cite{lee2023pix2struct}, a versatile patching strategy that maps visual elements to their corresponding HTML. This approach enables the model to adapt its image understanding capabilities to the specific context of UI-related screenshots.

\subsubsection{Task-Specific Fine-tuning.}
Although WebAgents can effectively understand environmental information through GUI comprehension training, accurately reasoning and generating the next steps for web interactions based on user tasks remains a significant challenge due to the complexity of web environments and the diversity of user objectives in web automation. 
Therefore, task-specific fine-tuning aims to equip WebAgents with web task-oriented skills, such as planning, reasoning, and interacting capabilities~\cite{shen2024falcon,furuta2023exposing,xu2024aguvis,guan2023intelligent}.
For instance,
\citet{gur2023real} propose an LLM-driven agent fine-tuned with scripted planning datasets, which decomposes natural language instructions into manageable sub-instructions, summarizes lengthy HTML documents into task-relevant snippets, and executes actions through self-generated Python programming. 
LCoW~\cite{lee2025learning} enhances WebAgents’ decision-making capabilities by leveraging contextualized observations from a Contextualization Module.
Specifically, this module is iteratively fine-tuned by the optimal contextualized observations sampling from its own outputs.
Besides, to efficiently navigate in the exponentially large exploration space, NNetNav~\cite{murty2024nnetscape} utilizes the hierarchical structure of language instructions for fine-tuning, rendering the search process more manageable. 
It prunes interaction episodes automatically when an intermediate trajectory cannot be meaningfully annotated, optimizing exploration effectiveness.
Besides, WebGUM~\cite{furuta2023multimodal} proposes a data-driven offline supervised fine-tuning process that adapts WebAgents for web navigation tasks based on instruction-following format problems and chain-of-thought examples across a variety of domains.
% WebGUM~\cite{furuta2023multimodal} transforms web navigation tasks into the instruction-following task that enables WebAgents to perform a data-driven offline supervised fine-tuning paradigm with instruction-following format problems and chain-of-thought examples across a variety of domains.
The web navigation action output is in free-form text, enhancing the flexibility and adaptability to tackle real-world web navigation.
In addition, LLMPA~\cite{guan2023intelligent} implements an end-to-end finetuning framework to accomplish multi-step operations based on high-level user command by integrating various auxiliary modules, like instruction decomposition and action prediction.
Notably, apart from these key supporting modules, LLMPA employs Controllable Calibration to meticulously evaluate predicted actions, effectively mitigating the hallucination issues prevalent in LLMs and enhancing overall execution reliability.

\subsubsection{Post-training.}
Post-training allows WebAgents to continuously adapt and improve after supervised training when facing exponentially large and dynamic web environments.
Given the open-domain nature of interactive web exploration, training methods relying solely on static datasets face clear limitations, making reinforcement learning~\cite{kaelbling1996reinforcement} a key post-training technique.
As web interfaces constantly evolve and user requirements change, reinforcement learning enables WebAgents to adapt in real-time by exploring dynamic environments and learning from interaction feedback~\cite{fereidouni2024grounded,humphreys2022data,thil2024navigating}.
For instance,
AutoGLM~\cite{liu2024autoglm} follows a progressive reinforcement learning framework that enables a continuous self-evolving learning paradigm through autonomous interactions with web environments.
The knowledge it acquires from real websites is inherently dynamic, enabling the system to adapt and refine its decision-making capability in real-time.
Moreover, reinforcement learning can further enhance the model’s capabilities by building upon the strengths of previous training phases.
AutoWebGLM~\cite{lai2024autowebglm} employs a multi-stage training approach to sequentially integrate different training strategies that significantly enhance its proficiency in planning, reasoning, and interacting.
To be specific, AutoWebGLM begins with fine-tuning to establish fundamental capabilities to fulfill user commands, followed by self-sampling reinforcement learning to mitigate hallucination caused by neglecting important states or operations.
RUIG~\cite{zhang2023reinforced} supervises the token sequence guided by visually semantic metrics during reinforcement learning to enhance GUI understanding capabilities derived from the pixel-to-sequence learning paradigm during prior training.

\section{Trustworthy WebAgents}\label{sec:trustworthy}

In recent years, the rapid evolution of LFM-empowered agents has ushered in a new era, profoundly reshaping various facets of web applications, including E-commerce, healthcare, and education. However, alongside this remarkable progress, a growing body of research and real-world incidents has highlighted the inherent risks and challenges associated with WebAgents~\cite{yang2025watch,schwartz2023enhancing,deng2024ai}. Recent studies have begun to reveal the potential threats posed by agents closely integrated with web systems. These threats range from unreliable and opaque decision-making in safety-critical contexts to the perpetuation of bias and unfairness towards marginalized communities~\cite{deng2024ai,li2024personal}.
Moreover, concerns have been raised about the inadvertent exposure of user privacy and sensitive business information, further emphasizing the critical need to ensure the security of WebAgent technologies.
Additionally, limitations in generalizability pose serious risks for WebAgents when confronted with diverse and unforeseen situations, potentially leading to critical failures in handling out-of-distribution data or operating across different domains.
Consequently, developing advanced techniques for trustworthy WebAgents has become an area of increasing interest for researchers, primarily focusing on creating and deploying AI systems that prioritize \textbf{Safety \& Robustness}, \textbf{Privacy}, and \textbf{Generalizability}.

\subsection{Safety \& Robustness}
WebAgents need to be resilient to both noisy changes and adversarial attacks, which is essential due to the complexity and volatility of the real web environments in which they operate~\cite{yuan2024r,guo2025redcode,boisvert2025doomarena}.
Motivated by these concerns, AdvWeb~\cite{xu2024advweb} explores the vulnerabilities of VLM-based WebAgents to black-box attacks, such as the injection of adversarial prompts into webpages. These vulnerabilities could result in significant issues, including inappropriate stock purchases or erroneous bank transactions.
Another instance of such threats is WIPI, a novel web threat that indirectly manipulates WebAgents to carry out malicious instructions embedded in publicly accessible webpages~\cite{wu2024wipi}.
To provide a thorough investigation, several rigorous benchmarks have been presented.
For example, ARE~\cite{wudissecting} includes 200 targeted adversarial tasks and assessment scripts within a realistic threat model using VisualWebArena~\cite{koh2024visualwebarena}, a real-world environment for WebAgents. 
ST-WebAgentBench~\cite{levy2024st} presents an online benchmark tailored to assess the safety and trustworthiness of WebAgents within enterprise environments.
RedCode~\cite{guo2025redcode} offers an evaluation platform complete with benchmarks for AI-assisted coding and software development, which is based on four fundamental principles: genuine interaction with systems, comprehensive assessment of unsafe code generation and execution, a variety of input formats, and high-quality safety scenarios and tests.
Additionally, \citet{kumar2024refusal} develop a comprehensive test suite called BrowserART, specifically designed for red-teaming browser agents. This suite comprises 100 diverse harmful behaviors related to browsers. Their findings indicate that refusal-trained models can be easily compromised when functioning as browser agents.

Beyond security investigations of WebAgents, advanced methods have been proposed to address the safety concerns.  
For example, Step~\cite{sodhi2023step} dynamically composes policies for web actions into a Markov Decision Process, ensuring that WebAgents effectively manage the handoff of control between different policies.
Another common solution involves enabling WebAgents to converse with each other to successfully accomplish complex tasks~\cite{wu2023autogen}.
This collaborative approach allows agents to verify the accuracy and appropriateness of actions before execution.
More recently, a novel method called Robust TBAS (RTBAS)~\cite{zhong2025rtbas} has been proposed, which selectively propagates security metadata in WebAgent tool calls using two novel screeners: LM-Judge Screening and Attention-Based Screening. These screeners are designed to identify relevant regions for generating the next response or tool call, while irrelevant regions are masked and redacted from the history.

\subsection{Privacy}
WebAgents have demonstrated remarkable potential in autonomously completing a wide range of tasks on real websites, significantly boosting human productivity.
However, web tasks, such as booking flights, often involve users' personal data, financial details, and proprietary business information, which may be exposed to potential privacy risks if WebAgents accidentally interact with compromised websites~\cite{zhang2024privacyasst}.
Protecting privacy is crucial to safeguarding the data used by WebAgents from breaches and unauthorized access.
Building on this understanding, \citet{wang2025unveiling} have explored the privacy risks associated with LLM Agents, particularly concerning their memory.
Their research introduces a black-box attack known as the Memory EXTRaction Attack (MEXTRA), which examines the vulnerability of LLM agents in preventing the extraction of private information from memory across various scenarios, including web applications.
Furthermore, \citet{liao2024eia} propose an Environmental Injection Attack (EIA) to explore two adversarial objectives within web environments: stealing specific personal information from users or capturing entire user requests.
It involves injecting malicious content that is tailored to adapt effectively to the environments in which the agents operate.
Additionally, PrivacyLens~\cite{shao2025privacylens} introduces a new dataset designed to assess the privacy norm awareness of language model agents in personalized communication scenarios, such as sending emails and writing social media posts.
Utilizing the PrivacyLens dataset, which includes a collection of privacy norms, the authors expand privacy-sensitive seeds into detailed vignettes and further into agent trajectories. This approach enables a multi-level evaluation of privacy leakage in the actions of language model agents.

\subsection{Generalizability}
The effectiveness of many WebAgents depends on the assumption that both their training and testing data originate from the same distribution. However, this assumption often fails due to the out-of-distribution (OOD) issue, which occurs when the training and testing data come from different distributions~\cite{li2025effects,patel2024large,bai2025digirl}. This mismatch poses significant challenges for deploying WebAgents in critical situations, highlighting the need to ensure generalizability in trustworthy WebAgents.
To address this challenge, Mind2Web~\cite{deng2023mind2web} has been developed as the first dataset aimed at creating and evaluating generalist WebAgents. 
Mind2Web includes over 2,000 open-ended tasks gathered from 137 websites across 31 domains, along with crowd-sourced action sequences for these tasks. 
It offers three essential components for building generalist WebAgents: (i) a variety of domains, websites, and tasks, (ii) the use of real-world websites rather than simulated or simplified ones, and (iii) a wide range of user interaction patterns.
Furthermore, \citet{patel2024large} investigate the ability of LLMs to enhance their performance autonomously as agents, which involves fine-tuning on data generated by the models themselves, particularly in long-horizon tasks within a complex environment, using the WebArena benchmark.
Recently, a World-Model-Augmented (WMA) WebAgent has been introduced~\cite{chae2024web}, representing a groundbreaking advancement in incorporating world models into LFM-empowered WebAgents.
WMA lays the groundwork for policy adaptation by utilizing feedback from simulated environments during web navigation, and its LLM backbone is trained with a novel method known as transition-focused observation abstraction.
Moreover, a novel autonomous reinforcement learning (RL) approach, called DigiRL~\cite{bai2025digirl}, is proposed for training in-the-wild device control agents. This approach involves fine-tuning a pre-trained VLM in two stages: offline and offline-to-online RL.
% Additionally, using reinforcement learning techniques, 
Additionally, \citet{mazzaglia2025genrl} introduce an agent learning framework called GenRL, which connects and aligns the representation of foundation VLMs with the latent space of generative world models for RL without any language annotations.
This framework enables the specification of tasks through vision and/or language prompts, grounds these tasks in the dynamics of the embodied domain, and facilitates the learning of corresponding behaviors through imagination.

In addition to the three dimensions previously discussed, trustworthy WebAgents also encompass other important dimensions, such as Fairness and Explainability~\cite{liu2022trustworthy}. While these aspects are crucial, they are still in the early stages of development, with limited literature available, particularly concerning WebAgent techniques. Therefore, we will explore these dimensions as future research directions requiring dedicated efforts in the next section.

\section{Future Directions}

\label{sec:future}
As research on WebAgents remains in its nascent stages, this section aims to discuss several promising research directions in the future. 

\subsection{Trustworthy WebAgents}
Existing research~\cite{aird2024dynamic} primarily focuses on enhancing the capabilities of WebAgents, while their trustworthiness remains largely under-explored, particularly in terms of Fairness and  Explainability~\cite{wang2024ali,liu2022trustworthy}. 
\textbf{Fairness} requires WebAgents to operate without bias in perception, reasoning, and execution. For instance, when users of different genders request WebAgents to search for suitable job positions, the agents should handle these requests fairly rather than relying on prior assumptions, such as suggesting that men are more suited to be lawyers while women are more suited to be nurses. 
\textbf{Explainability} requires that WebAgents be capable of justifying their actions, helping users understand their internal mechanisms, and ensuring their reliability in high-stakes environments, such as stock investment~\cite{abdelnabi2024cooperation} and molecular design~\cite{liu2024moleculargpt}.
Furthermore, efforts can be devoted to studying other aspects of achieving trustworthy WebAgents, such as safety\&robustness and privacy.

\subsection{Datasets and Benchmarks of WebAgents}
The systematic evaluation of WebAgents has emerged as a critical research focus, with several comprehensive benchmarks developed to rigorously assess their performance~\cite{xu2024tur, shlomov2024grounding, zhou2023webarena, drouin2024workarena, boisvert2024workarena++, debrowsergym}. 
For instance, PersonalWAB~\cite{cai2024large} provides a comprehensive benchmark encompassing 1,000 diverse user-profiles and over 40,000 web behaviors derived from real-world data.
It incorporates user instructions, personalized user data, web functions, and two evaluation paradigms (i.e., profile-behavior consistency and profile-product consistency) across various personalized web tasks.
Additionally, the Webcanvas~\cite{pan2024webcanvas} framework has emerged as a notable tool specifically designed to evaluate WebAgents in real-time online environments. 
Webcanvas simulates realistic web-based interactions, allowing researchers to observe how agents perform in dynamic, unpredictable settings that mirror user experiences.
These benchmarks test the agents’ proficiency in web navigation, task automation, and interaction with dynamic online interfaces, providing valuable insights into their strengths and limitations. 
Despite the remarkable success, most benchmarks focus on specific, limited facets of functionality or particular use cases, often overlooking crucial aspects such as adapting to varied web layouts, resilience in the face of unforeseen errors, or proficiency in managing intricate, multi-faceted tasks. 
Moreover, many existing assessments fail to adequately address real-world complexities, including fluctuating internet speeds, inconsistent website structures, and the need for sustained contextual reasoning across prolonged interactions. 
Therefore, there is an urgent need for more comprehensive and well-rounded benchmarks that can fully evaluate the wide-ranging capabilities of WebAgents.

\subsection{Personalized WebAgents}
While existing LFM-empowered WebAgents have demonstrated impressive capabilities in various web tasks, they still face significant limitations in personalization due to their massive parameter size (often in the billions) and expensive training costs.
This limitation often results in generic responses or actions that fail to align with individual users' unique needs or expectations, significantly reducing user experience and satisfaction. 
Therefore, developing WebAgents that simultaneously demonstrate robust capabilities while offering meaningful personalization represents a challenging yet promising research direction.
To tackle this challenge, researchers have begun exploring innovative approaches, such as integrating Retrieval-Augmented Generation (RAG) systems ~\cite{fan2024survey,ni2025towards} with memory mechanisms that encompass both long-term~\cite{zhong2024memorybank, gutierrez2025rag} and short-term~\cite{mukherjee2025documentsdialoguebuildingkgrag, zhong2024design} components. 
Specifically, the long-term memory supports WebAgents by retrieving and storing relevant information over extended periods, enabling WebAgents to maintain a consistent and personalized understanding of the user.
In contrast, short-term memory allows WebAgents to quickly adapt to immediate contextual cues in real-time prompts, ensuring responsive and dynamic adjustments to the current conversation or task.
By integrating these elements, personalized WebAgents can achieve higher adaptability, offering tailored solutions that better meet individual user needs.

\subsection{Domain-Specific WebAgents}
Recent studies have increasingly demonstrated the promising applications of LFM-empowered WebAgents across various vertical domains~\cite{li2024agent}.
However, the application of LFM-empowered WebAgents in specialized professional domains such as education and healthcare remains largely unexplored despite the urgent needs and significant potential benefits in these fields. 
Consequently, adapting general-purpose WebAgents to domain-specific contexts has emerged as a promising research direction. 
To achieve reliable and powerful domain-specific WebAgents, 
they must be equipped with a customized knowledge base reflecting the nuances of the domains, a robust mechanism for securely managing sensitive data, and the flexibility to adapt to the needs of rapidly evolving industries. These advancements open up numerous research opportunities to improve and enhance WebAgents’ capabilities.
\section{Conclusion}
Web tasks have become an integral part of people's daily lives, as individuals rely on the web for various activities. 
Despite the importance of the web, many daily web tasks remain repetitive and time-consuming. 
Recently, extensive efforts have been made to develop autonomous WebAgents capable of completing these complex web tasks based on user instructions. 
With the rapid advancements in WebAgents, there is an urgent need to review existing studies systematically. 
To address this gap, in this paper, we comprehensively summarize recent developments in WebAgents from three perspectives: architectures, training, and trustworthiness, providing researchers with a clear overview of this field.  
Furthermore, as research on WebAgents is still in its early stages, we discuss potential research directions to offer insights and inspiration to the research community.

% \begin{acks}
% \end{acks}

\balance
\bibliographystyle{ACM-Reference-Format}
\bibliography{Reference}

% \appendix

\end{document}